\documentclass[10pt,twocolumn,letterpaper]{article}

\usepackage{cvpr}              
\usepackage{amsmath,amssymb}
\usepackage{algorithm}
\usepackage{algpseudocode}
\usepackage{multirow}

\definecolor{cvprblue}{rgb}{0.21,0.49,0.74}
\usepackage[pagebackref,breaklinks,colorlinks,allcolors=cvprblue]{hyperref}
\usepackage{color}

\def\paperID{7998} 
\def\confName{CVPR}
\def\confYear{2026}

\title{StyleTextGen: Style-Conditioned Multilingual Scene Text Generation}
\author{
Zeyu Chen\textsuperscript{1} \quad
Fangmin Zhao\textsuperscript{3} \quad
Yan Shu\textsuperscript{2} \quad
Yichao Liu\textsuperscript{1} \quad
Liu Yu\textsuperscript{1} \quad
Yu Zhou\textsuperscript{1}\thanks{Corresponding author.} \\
\textsuperscript{1}Nankai University \quad
\textsuperscript{2}University of Trento\\
\textsuperscript{3}Institute of Information Engineering, Chinese Academy of Sciences\\
}

\begin{document}
\maketitle

\begin{abstract}
Style-conditioned scene text generation faces unique challenges in extracting precise text styles from complex backgrounds and maintaining fine-grained style consistency across characters, especially for multilingual scripts. We propose StyleTextGen, a novel framework that learns to perceive and replicate visual text styles across different languages and writing systems. Our approach features three key contributions: 
First, we introduce a dual-branch style encoder dedicated to style modeling, yielding robust multilingual text style representations in complex real-world scenes. 
Second, we design a text style consistency loss that enhances style coherence and improves overall visual quality. 
Third, we develop a mask-guided inference strategy that ensures precise style alignment between generated and reference text. 
To facilitate systematic evaluation, we construct StyleText-CE, a bilingual scene text style benchmark covering both monolingual and cross-lingual settings.
Extensive experiments demonstrate that StyleTextGen significantly outperforms existing methods in style consistency and cross-lingual generalization, establishing new state-of-the-art performance in multilingual style-conditioned text generation.
\end{abstract}    
\section{Introduction}
\label{sec:intro}

\begin{figure}[ht]
\centering
\includegraphics[width=0.9\columnwidth]{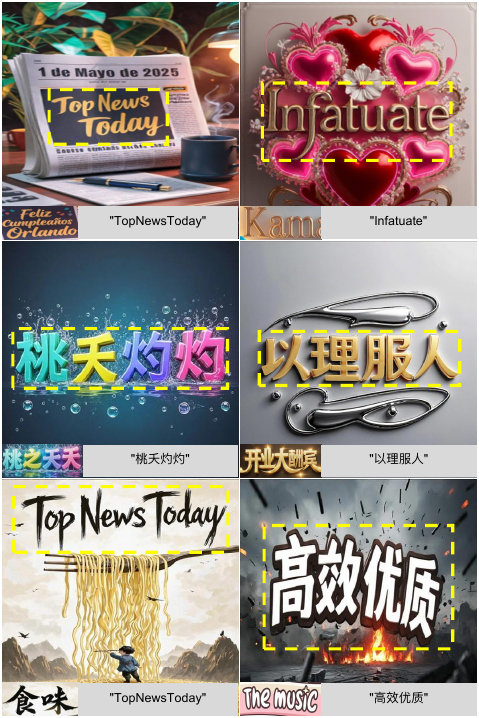}
\caption{Examples of our StyleTextGen for style-conditioned scene text generation. 
Left: self-style reference (in-place generation). Right: external-style reference. 
For each example, the reference style image and the generated target text are shown at the bottom of the panel. }
\label{fig:res}
\end{figure}

Text-to-image generation \cite{Dustin_SDXL_ICLR24,flux2024} has witnessed remarkable progress in synthesizing photorealistic images from textual descriptions. However, generating scene text with precise style control remains a fundamental challenge. 
The task of style-conditioned scene text generation aims to synthesize text in images while faithfully replicating the visual style from reference example, including font characteristics, stroke pattern, color gradient and texture detail. 
This capability is crucial for real-world applications such as multilingual signage design and advertising poster creation, and also benefit scene text detection, recognition, and spotting \cite{zheng2024cdistnet,du2025instruction,du2025context,yang2025ipad,lyu2025arbitrary,cao2025devil}.
Unlike semantic content that can be specified through text prompts, these nuanced visual attributes require visual style references, making image-conditioned generation a natural solution for this task.

Despite its importance, style-conditioned scene text generation faces unique technical challenges that distinguish it from general image synthesis. First, extracting text-specific style features from complex scene images is non-trivial—reference images often contain cluttered backgrounds, varying lighting conditions, and diverse text layouts that can confound style perception. Existing scene text editing methods \cite{zeng2024textctrl,fang2025recognition} typically reuse styles from the original in-place text, limiting their ability to adopt external style references. Recent advances such as Calligrapher~\cite{ma2025calligrapher} attempt free-style text generation, but their style extraction mechanisms lack explicit content awareness, leading to style-content entanglement and degraded performance when processing complex backgrounds. Second, maintaining fine-grained style consistency across all characters becomes substantially more challenging in multilingual scenarios, where different writing systems (e.g., Latin, Chinese, Arabic) exhibit distinct typographic structures and varying numbers of strokes. The lack of robust cross-lingual style transfer mechanisms in current methods results in inconsistent character appearance and structural distortions when generating text in languages different from the reference style.

To address these challenges, we propose StyleTextGen, a novel style-conditioned framework for multilingual scene text generation.  First, we design a dual-branch style encoder 
dedicated to modeling text style. 
A text-focused branch pretrained on multilingual text segmentation captures fine-grained typographic features, while a visual prior branch extracts holistic scene information. Their fusion yields robust, text-aware style embeddings that remain stable across complex backgrounds and diverse languages. Second, we propose a text style consistency loss that enforces style alignment specifically within text regions. This mask-guided loss promotes uniform style attributes across all generated characters, significantly improving visual coherence in multilingual scenarios. Third, we introduce a mask-guided inference strategy that extracts style features from reference images through mask-based inversion and injects them into the generation process. This refinement ensures precise style correspondence between reference and generated text across different writing systems.

To enable systematic evaluation, we construct StyleText-CE, a bilingual benchmark supporting four settings: self-style reference, external-style reference, monolingual generation, and cross-lingual transfer. As shown in Figure~\ref{fig:res}, StyleTextGen accurately transfers text styles while maintaining readability across diverse languages and styles.

Our contributions are summarized as follows:

\begin{itemize}
\item We propose a dual-branch style encoder that robustly captures style features from multilingual text, combined with a text style consistency loss to enhance style coherence and visual consistency.
\item We design an inference strategy that enhances style alignment between generated scene text and reference, improving fine-grained style consistency across characters.
\item We construct a bilingual scene text style benchmark with monolingual and cross-lingual settings, demonstrating superior performance in style consistency and
 cross-lingual generalization.
\end{itemize}

\section{Related Work}

\subsection{Scene Text Generation}

Traditional text-to-image diffusion models~\cite{Saharia_Imagen_NIPS22, DALLE3, Rombach_LDM_CVPR22, Ramesh_DALLE_icml21, shu2025visual, shu2025semantics, yang2025vidtext}, struggle to render legible and semantically correct text in complex scenes. This limitation has motivated a line of work on dedicated scene text generation. Methods such as DiffSTE~\cite{diffste}, UDiffText~\cite{zhao_udifftex}, GlyphByT5~\cite{glyphbyt5}, and SceneTextGen~\cite{zhangli2024layoutagnosticscenetext} improve textual representation through character-level encoders. Apart from these, AnyText~\cite{tuo2023anytext}, GlyphDraw2~\cite{ma2025glyphdraw2}, GlyphOnly~\cite{li2024first}, TextDiffuser~\cite{Chen_TextDiffuser_Corr23}, TextDiffuser2~\cite{Chen_TextDiffuser2}, and GlyphControl~\cite{Yang_GlyphControl_Corr23} leverage glyph priors and layout masks to enhance structural fidelity and spatial controllability. 

Focusing more specifically on style-conditioned scene text generation, GlyphByT5~\cite{glyphbyt5} models partial font and color information within its text encoder and pre-rendered glyph space, enabling coarse-grained control over the appearance of generated text. AnyText2~\cite{tuo2024anytext2} further introduces explicit font and color encoders to provide additional style-related guidance during rendering. However, these approaches are typically restricted to relatively simple fonts and often struggle to synthesize artistic typography or faithfully follow a given style reference. 
To address these limitations, we propose StyleTextGen, which generates scene text conditioned on arbitrary text-style reference images.

\subsection{Scene Text Editing}

Scene text editing aims to modify the textual content in an image while preserving the background and maintaining style consistency. Many methods explicitly separate foreground text from background. SRNet~\cite{wu2019editing} and STEFANN~\cite{roy2020stefann} use dedicated modules for background inpainting and text re-rendering. SwapText~\cite{yang2020swaptext} extends SRNet with a thin-plate spline interpolation network to handle curved text. TextStyleBrush~\cite{krishnan2023textstylebrush} adopts a StyleGAN2~\cite{karras2020analyzing} based self-supervised framework to learn transferable text styles, and MOSTEL~\cite{qu2023exploring} employs a semi-supervised training scheme to improve robustness in the wild. STEEM~\cite{zhou2024explicitly} minimizes background reconstruction error to explicitly decouple style and content, leading to higher editing fidelity. Recent method TextCtrl~\cite{zeng2024textctrl} pre-train a style encoder to enforce style consistency, while GlyphMastero\cite{wang2025glyphmastero} fuses local character-level features with global text-line structures for editing complex Chinese characters. 
Overall, these methods focus on preserving background information and modifying text content. Although some of them consider text style, they cannot generate text that follows arbitrary external style references. StyleTextGen addresses this limitation by enabling scene text generation conditioned on any given style reference image.

\subsection{Image style transfer}

Image style transfer aims to apply the style of a reference image to a content image. Early neural style transfer methods~\cite{gatys2016image} formulate this problem as matching deep feature statistics, and subsequent feed-forward approaches~\cite{johnson2016perceptual, ulyanov2016texture} enable efficient fast stylization. AdaIN~\cite{huang2017arbitrary} supports arbitrary style transfer by aligning feature statistics, while style-based generative models such as StyleGAN~\cite{karras2019style, karras2020analyzing} further improve style controllability. More recent work leverages vision–language models and diffusion models for style control: CLIPstyler~\cite{kwon2022clipstyler} builds on CLIP~\cite{radford2021learning} to achieve text-guided stylization, 
diffusion-based methods enable style transfer guided by text or reference images
~\cite{brooks2023instructpix2pix, zhang2023inversion, lei2025stylestudio}, and IP-Adapter~\cite{ip-adapter} injects image-conditional style control into diffusion backbones. 
These methods mainly target holistic image stylization and are not designed for fine-grained rendering of local text regions with complex glyph structures. In contrast, StyleTextGen focuses on scene text generation conditioned on arbitrary text-style reference images, and provides explicit control over the style of the rendered text.

\begin{figure*}[t]
    \centering
    \includegraphics[width=0.87\textwidth]{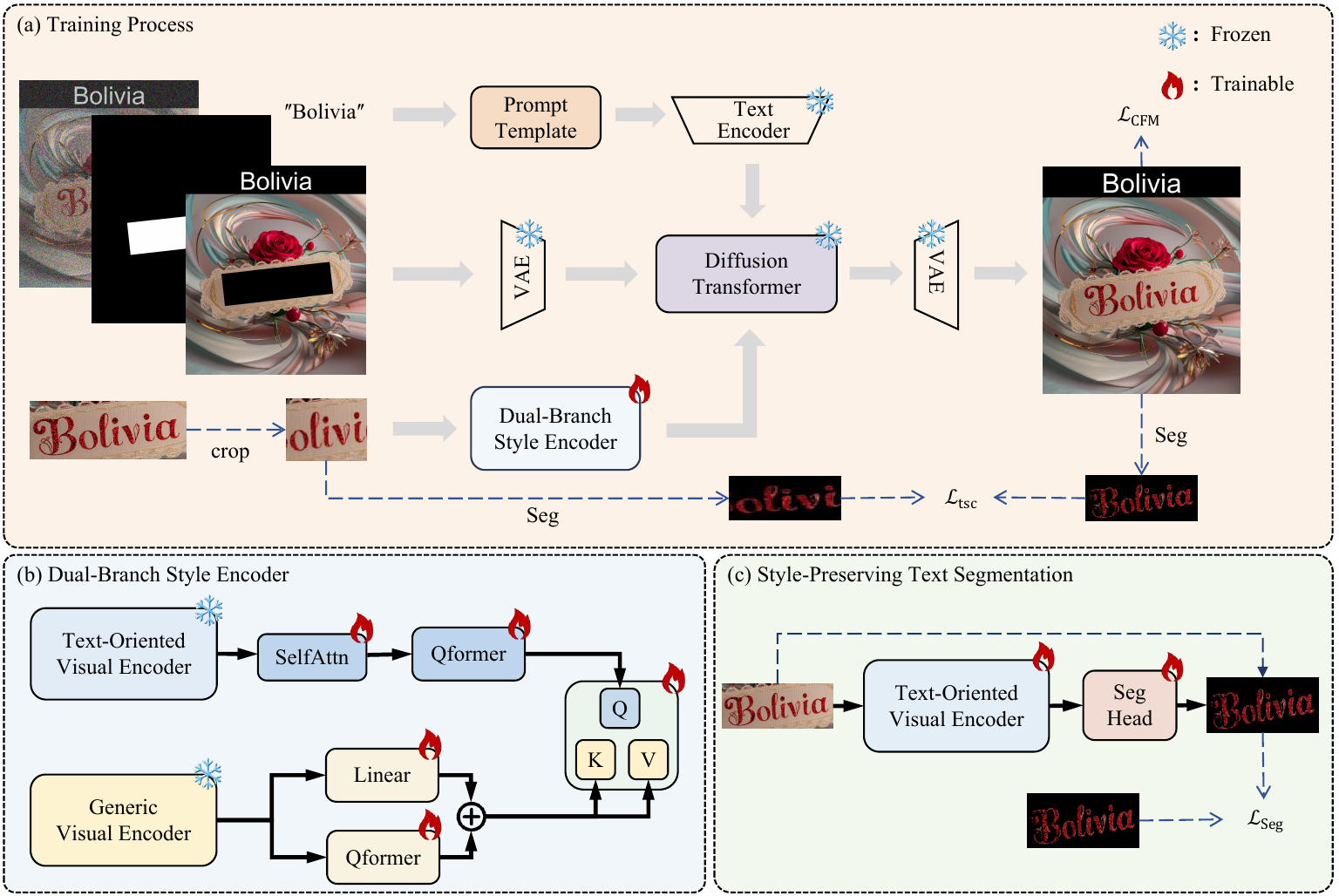}
    \caption{Overview of StyleTextGen. 
(a) Training process. The inpainting input to the diffusion transformer is constructed from a scene text image concatenated with its glyph map. A randomly cropped text patch serves as the style reference. The model is optimized with the flow-matching loss $\mathcal{L}_{\text{CFM}}$ on the inpainting task and the text style consistency loss $\mathcal{L}_{\text{tsc}}$ computed between the segmented reference and generated text regions.
(b) Dual-Branch Style Encoder. The style image is encoded by a text-oriented visual encoder and a generic visual encoder, whose outputs are combined to obtain a text-aware style embedding.
(c) Style-Preserving Text Segmentation. This pretraining task targets the text-oriented encoder and preserves text style and layout instead of producing binary masks, encouraging the encoder to learn rich style-aware text features.
}
    \label{fig:combined}
\end{figure*}

\section{Method}

\subsection{Preliminaries}

Recent advances in diffusion-based inpainting frameworks, such as FLUX.1-Fill-dev~\cite{flux2024} and BrushNet~\cite{ju2024brushnet}, demonstrate strong potential for context-aware image synthesis.
Building upon this line of work, we formulate our style-conditioned multilingual scene text generation model under an inpainting paradigm, similar to TextFlux, enabling controllable and high-fidelity transfer of visual text styles from reference images. First, we extract the target text from the given text prompt $p$, render it into a binary image $I_{\text{glyph}}$, and concatenate vertically it with the scene image $I_{\text{scene}}$ to provide explicit structural and style cues for text rendering:
\begin{equation}
I_{\text{concat}} =
\begin{bmatrix}
I_{\text{glyph}} \\
I_{\text{scene}}
\end{bmatrix},
\label{eq:concat}
\end{equation}

where $I_{\text{glyph}}\in\mathbb{R}^{H\times W\times1}$ encodes the textual structure and $I_{\text{scene}}\in\mathbb{R}^{H\times W\times3}$ provides the background context.

Next, with the inpainting mask $M$ that indicates the text generation position and a style reference image $I_{\text{style}}$ that provides style information, we define the conditioning set as $c=\{I_{\text{concat}}, M, p, I_{\text{style}}\}$.
Therefore, the overall architecture can be formally expressed as: $\hat{I}_{target}=\mathcal{D}(\mathcal{F}(\epsilon, c))$, herein, $\epsilon$ represents Gaussian noise, $\mathcal{D}$ denotes the VAE decoder, and $\mathcal{F}$ stands for the denoising process of DiT.
Following the rectified flow~\cite{wang2024rectifieddiffusion} view, let $x_0=\mathcal{E}(I_{\text{target}})$ denote the clean latent of the target image, and $\epsilon \sim \mathcal{N}(0,I)$ denote a standard Gaussian noise sample.
The forward process adds the noise to $x_0$ to generate the noisy sample $x_t$ with the linear noise adding schedule $\sigma_t$ at time-step $t$:
\begin{equation}
x_t = (1 - \sigma_t)x_0 + \sigma_t \epsilon
\label{eq:interp}.
\end{equation}
The DiT backbone predicts a conditional velocity field $v_{\theta}(x_t, t, c)$ and is optimized using the conditional flow-matching objective:
\begin{equation}
\mathcal{L}_{\text{CFM}} =
\mathbb{E}_{t,\,x_0,\,\epsilon}
\Big[
\omega_t
\big\|
v_{\theta}(x_t, t, c)
- (\epsilon - x_0)
\big\|_2^2
\Big]
\label{eq:cfm},
\end{equation}
where $\omega_t$ is a time-dependent weighting term. 

\subsection{Overall Architecture}

The overall architecture of our proposed StyleTextGen is shown in Figure \ref{fig:combined}. Given a background scene image, a text prompt and a style reference text image, StyleTextGen could render text within a designated region such that the generated text faithfully inherits the font, color, and texture characteristics of the reference style. In Section~\ref{subsec:style_encoder}, we present a dual-branch style encoder to tackle the challenge of extracting text styles from complex multilingual scenes, and further explain the injection of text-aware style embeddings for controllable style transfer.
To ensure style consistency between the generated text and the reference, we further introduce a text style consistency loss, whose formulation is described in Section~\ref{subsec:style_loss}. Furthermore, we propose an inference-time style injection strategy described in Section~\ref{subsec:style_align}, which refines the style consistency of the generated text by adjusting attention responses.

\subsection{Dual-Branch Style Encoder}
\label{subsec:style_encoder}

To robustly encode fine-grained text style cues under complex multilingual scenes, we design a \textbf{Dual‑Branch Style Encoder}. 
Given a style reference image $I_{\text{style}}$, the encoder consists of two complementary branches—a \emph{textual style branch} and a \emph{visual prior branch}. Their fused outputs yield a text‑aware style embedding $z_{\text{style}}$ that conditions the DiT backbone, with the detailed architecture illustrated in Figure~\ref{fig:combined}(b).

\textbf{Textual style branch.}
This branch captures text‑specific style cues such as glyph structure, stroke texture, and color distribution.
It contains three sequential components: 
a text-oriented encoder $E_{\text{text}}$, 
a self‑attention refinement transformer $S_{\text{text}}$, 
and a Q‑Former~\cite{Li_BLIP2_Corr23} adaptor $Q_{\text{text}}$.
The text-oriented encoder is pretrained on a style‑preserving multilingual text segmentation task, as in Figure~\ref{fig:combined}(c), encouraging the extraction of stable textual style features from complex backgrounds and diverse scripts.
The encoding process is formulated as
\begin{equation}
h_{\text{text}} = Q_{\text{text}}\!\big(S_{\text{text}}\!\big(E_{\text{text}}(I_{\text{style}})\big)\big),
\label{eq:textual_branch}
\end{equation}
where $h_{\text{text}}$ denotes the refined textual style representation.

\textbf{Visual prior branch.}
Built upon a generic vision encoder instantiated with a pretrained vision‑language backbone~\cite{tschannen2025siglip}, 
this branch extracts holistic visual information to stabilize and generalize the learned style embedding.
It employs a visual encoder $E_{\text{vis}}$, 
an MLP projector $P_{\text{vis}}$, 
and a Q‑Former module $Q_{\text{vis}}$:
\begin{equation}
h_{\text{vis}} = P_{\text{vis}}(E_{\text{vis}}(I_{\text{style}})) + Q_{\text{vis}}(E_{\text{vis}}(I_{\text{style}})).
\label{eq:visual_branch}
\end{equation}
This visual prior provides global appearance and tonal coherence complementary to the text‑specific features.

The two representations are fused through cross‑attention, 
using $h_{\text{text}}$ as queries and $h_{\text{vis}}$ as keys and values:
\begin{equation}
z_{\text{style}} = \mathrm{Attn}(h_{\text{text}},\,h_{\text{vis}},\,h_{\text{vis}}),
\label{eq:fusion}
\end{equation}
producing a robust text‑aware style embedding.

Finally, The fused embedding $z_{\text{style}}$ is linearly projected to style key–value pairs $(K_s,V_s)$,
which modulate the DiT attention through an additive style‑attention branch:
\begin{equation}
F_{\text{style}} = \mathrm{SelfAttn}(Q,K,V) + \mathrm{StyleAttn}(Q,K_s,V_s),
\label{eq:style_inject}
\end{equation}
where $F_{\text{style}}$ integrates contextual dependencies from self‑attention with style‑specific modulation.

\subsection{Text Style Consistency Loss}
\label{subsec:style_loss}

Background clutter may obscure style cues, and the absence of an explicit text-region objective often leads to character style drift.
We therefore introduce a \textbf{Text Style Consistency Loss} that aligns the style statistics of generated text with those of the reference within text regions.

Following the neural style representation approach of Gatys et al.~\cite{gatys2016image}, 
we extract multi-layer features from a pretrained visual encoder $\phi$ to capture the style statistics of an image, 
and let $J$ be a set of its layers. 
For layer $j$, denote $\phi_j(x)\in\mathbb{R}^{C_j\times H_j\times W_j}$ and reshape it to $F_j(x)\in\mathbb{R}^{C_j\times N_j}$ with $N_j=H_jW_j$.
The Gram matrix, which measures inter-channel correlations and summarizes texture appearance, is computed as:
\begin{equation}
G_j^{\phi}(x)=\frac{1}{N_j}\,F_j(x)\,F_j(x)^{\top}.
\label{eq:gram}
\end{equation}
The text style consistency loss is then defined as the Frobenius distance between 
the Gram matrices of the text regions in the generated image and the reference. Given the generated image $\hat{x}$ with text mask $M_{\text{gen}}$ and the reference image $I_{style}$ with mask $M_{\text{ref}}$, the loss is
\begin{equation}
\mathcal{L}_{\text{tsc}}=
\sum_{j\in J}
\Big\|
G_j^{\phi}\!\big(M_{\text{gen}}\!\odot\!\hat{x}\big)
-
G_j^{\phi}\!\big(M_{\text{ref}}\!\odot\!I_{style}\big)
\Big\|_{F}^{2}.
\label{eq:tsc_loss}
\end{equation}
Finally, the overall training objective combines the conditional flow-matching loss and the text style consistency loss as:
\begin{equation}
\mathcal{L}=\mathcal{L}_{\text{CFM}}+\lambda_{\text{tsc}}\,\mathcal{L}_{\text{tsc}},
\label{eq:total_loss}
\end{equation}
where $\lambda_{\text{tsc}}$ balances the influence of the style consistency term and is set to 10 by default.

\subsection{Inference-time Style Injection}
\label{subsec:style_align}
In this section, we introduce an inference-time style injection strategy that directly refines the attention responses using the reference features.

Given a generated image $\hat{I}_{gen}$ and a style reference image $I_{style}$, 
we first employ a bilingual text segmentation model to extract the text-region masks for both images, 
denoted as $M_{\text{gen}}$ and $M_{\text{style}}$. 
These masks enable localized feature manipulation within text areas for subsequent style injection.

Next, we invert the latent feature $x_s$ of reference image $I_{style}$ through DiT to recover its internal key–value features $(K_s, V_s)$, 
where the mask $M_{\text{style}}$ ensures that only text regions contribute to the extracted style representation. 
This inversion~\cite{rout2024semantic,lipman2023flowmatching} effectively maps the reference image back to the model’s latent attention space, 
providing style-aware feature pairs for mask-guided modulation.

Specifically, we first obtain the style-adapted key and value features via AdaIN:
\begin{equation}
    \tilde{K}, \tilde{V} = \operatorname{AdaIN}(K, V; K_s \odot M_{\text{style}}, V_s \odot M_{\text{style}}),
\end{equation}
and then combined with the original ones via $M_{gen}$:
\begin{equation}
\begin{aligned}
K' &= (1 - M_{\text{gen}})\odot K + M_{\text{gen}}\odot \tilde{K},\\
V' &= (1 - M_{\text{gen}})\odot V + M_{\text{gen}}\odot \tilde{V}.
\end{aligned}
\label{eq:style_kv_update}
\end{equation}

Next, under the guidance of these masks, we accurately extract style-related features and ensure that such style control is precisely applied to the text regions of interest:

\begin{equation}
\begin{aligned}
f_{\text{base}} &= \operatorname{Attention}(Q, K', V'),\\
f_{\text{style}} &= \operatorname{Attention}(Q, K_s, V_s),\\
\end{aligned}
\label{eq:style_attn}
\end{equation}
and obtain the final output via mask-based blending:
\begin{equation}
\begin{aligned}
f_{out} &= (1 - M_{\text{gen}})\odot f_{\text{base}}  
+ \\&M_{\text{gen}}\odot \operatorname{AdaIN}(f_{\text{base}};\ f_{\text{style}} \odot M_{\text{style}}).
\end{aligned}
\label{eq:style_output}
\end{equation}

To preserve scene-text fidelity and maintain background coherence, 
we apply the style injection only during the first 10 denoising steps of the generation process.

\begin{table*}[ht]
\centering
\small
\setlength{\tabcolsep}{3pt}
\caption{Comparison with state-of-the-art multi-lingual (English and Chines) methods. $\uparrow$/$\downarrow$ indicates higher/lower is better. Our approach outperforms prior methods on all metrics.}
\begin{tabular}{@{}l|cccc|cccc@{}}
\toprule
\multirow{2}{*}{Model} & \multicolumn{4}{c|}{Text Accuracy} & \multicolumn{4}{c}{Style Similarity} \\
\cmidrule(lr){2-5} \cmidrule(lr){6-9}
 & \multicolumn{2}{c}{Sen.Acc $\uparrow$} & \multicolumn{2}{c|}{NED $\uparrow$} & \multicolumn{2}{c}{FID $\downarrow$} & \multicolumn{2}{c}{LPIPS $\downarrow$} \\
& English & Chinese & English & Chinese & English & Chinese & English & Chinese \\
\midrule
AnyText & 0.5613 & 0.4986 & 0.7264 & 0.6691 & 67.82 & 71.35 & 0.212 & 0.267 \\
Calligrapher & 0.6129 & 0.5153 & 0.7637 & 0.6847 & 54.26 & 63.28 & 0.176 & 0.231 \\
TextFlux & 0.6542 & 0.6185 & 0.8018 & 0.7934 & 59.44 & 57.12 & 0.187 & 0.204 \\
Ours & \textbf{0.7102} & \textbf{0.6524} & \textbf{0.8575} & \textbf{0.8187} & \textbf{49.51} & \textbf{52.43} & \textbf{0.161} & \textbf{0.187} \\
\bottomrule
\end{tabular}
\label{tab:model_comparison}
\end{table*}

\begin{figure}[ht]
  \centering
   \includegraphics[width=1.0\linewidth]{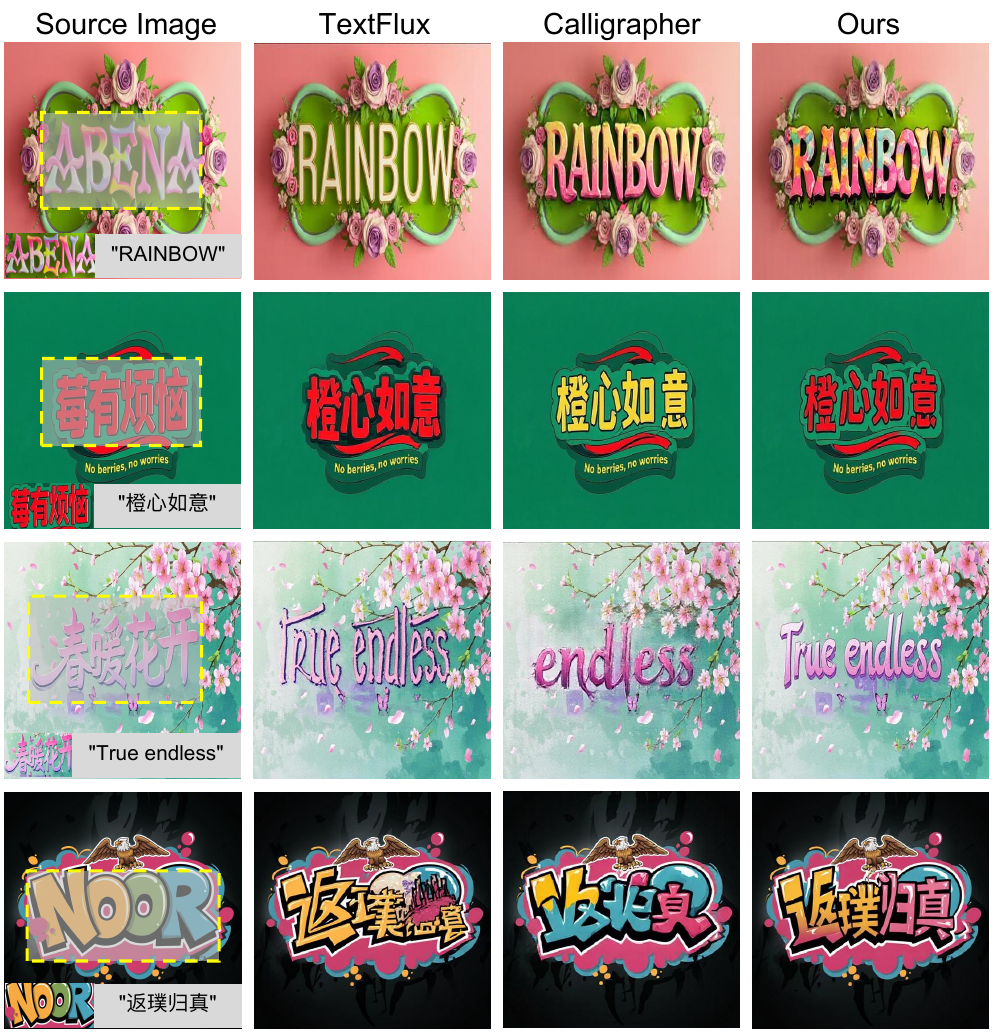}
   \caption{Qualitative comparison on the StyleText-CE benchmark under the self-style reference setting.}
   \label{fig:qua1}
\end{figure}

\begin{figure*}[tbp]
  \centering
   \includegraphics[width=1.0\linewidth]{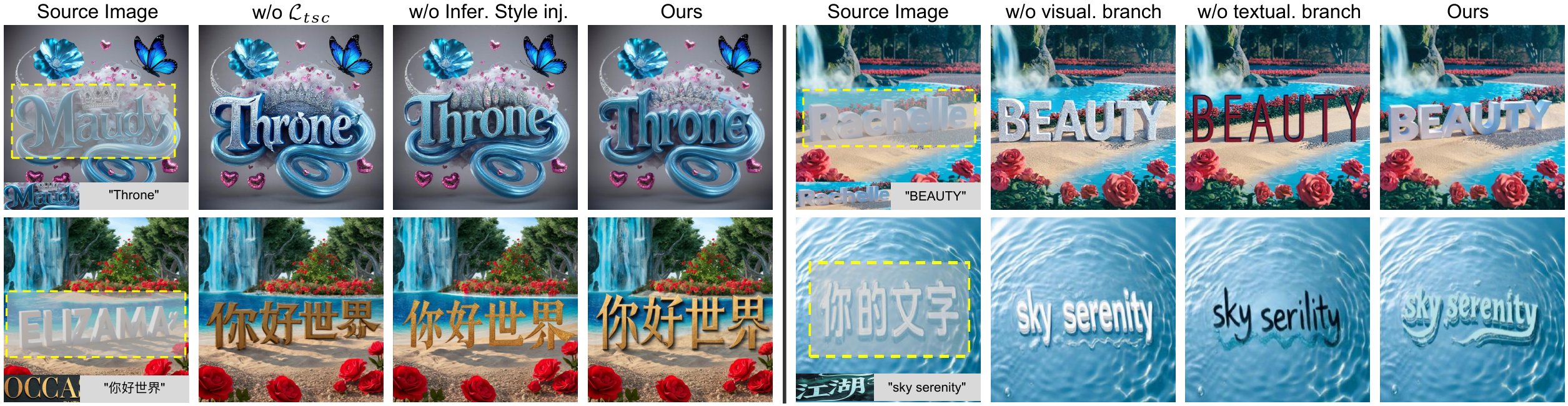}
   \caption{Qualitative results of the ablation study. The left group shows the effects of removing the Text Style Consistency Loss ($\mathcal{L}_{tsc}$) and the inference-time style injection module. The right group presents ablations of the Dual-Branch Style Encoder by removing either the visual prior branch or the textual style branch. The first row corresponds to the self-style reference setting, while the second row shows the external-style reference setting.}
   \label{fig:ab}
\end{figure*}

\begin{table*}[t]
\centering
\small
\setlength{\tabcolsep}{4pt}
\caption{Quantitative comparison between Calligrapher and StyleTextGen on the StyleText-CE benchmark under different style reference settings. Columns correspond to monolingual (cn, en) and cross-lingual (cn$\rightarrow$en, en$\rightarrow$cn) configurations.}
\begin{tabular}{@{}l|cccc|cccc@{}}
\toprule
\multirow{2}{*}{Model} & 
\multicolumn{4}{c|}{Self Style Reference} & 
\multicolumn{4}{c}{External Style Reference} \\
\cmidrule(lr){2-5} \cmidrule(lr){6-9}
& cn & en & cn$\rightarrow$en & en$\rightarrow$cn & cn & en & cn$\rightarrow$en & en$\rightarrow$cn \\
\midrule
\multicolumn{9}{l}{\textit{Text Accuracy (Sen.Acc $\uparrow$ / NED $\uparrow$)}} \\
Calligrapher & 0.51 / 0.67 & 0.64 / 0.78 & 0.57 / 0.74 & 0.49 / 0.66 & 0.49 / 0.67 & 0.58 / 0.75 & 0.55 / 0.72 & 0.47 / 0.64 \\
Ours & \textbf{0.66} / \textbf{0.81} & \textbf{0.70} / \textbf{0.85} & \textbf{0.66} / \textbf{0.80} & \textbf{0.64} / \textbf{0.77} & \textbf{0.60} / \textbf{0.79} & \textbf{0.71} / \textbf{0.84} & \textbf{0.67} / \textbf{0.77} & \textbf{0.65} / \textbf{0.82} \\
\midrule
\multicolumn{9}{l}{\textit{Style Similarity (FID $\downarrow$ / LPIPS $\downarrow$)}} \\
Calligrapher & 86.24 / 0.44 & 82.37 / 0.45 & 134.82 / 0.50 & 149.68 / 0.52 & 136.41 / 0.53 & 105.26 / 0.48 & 137.94 / 0.55 & 147.83 / 0.55 \\
Ours & \textbf{76.95} / \textbf{0.42} & \textbf{77.48} / \textbf{0.40} & \textbf{124.36} / \textbf{0.48} & \textbf{135.27} / \textbf{0.51} & \textbf{117.92} / \textbf{0.52} & \textbf{93.54} / \textbf{0.44} & \textbf{125.88} / \textbf{0.54} & \textbf{129.63} / \textbf{0.50} \\
\bottomrule
\end{tabular}
\label{tab:styletext_ce_results}
\end{table*}

\begin{figure}[ht]
  \centering
   \includegraphics[width=1.0\linewidth]{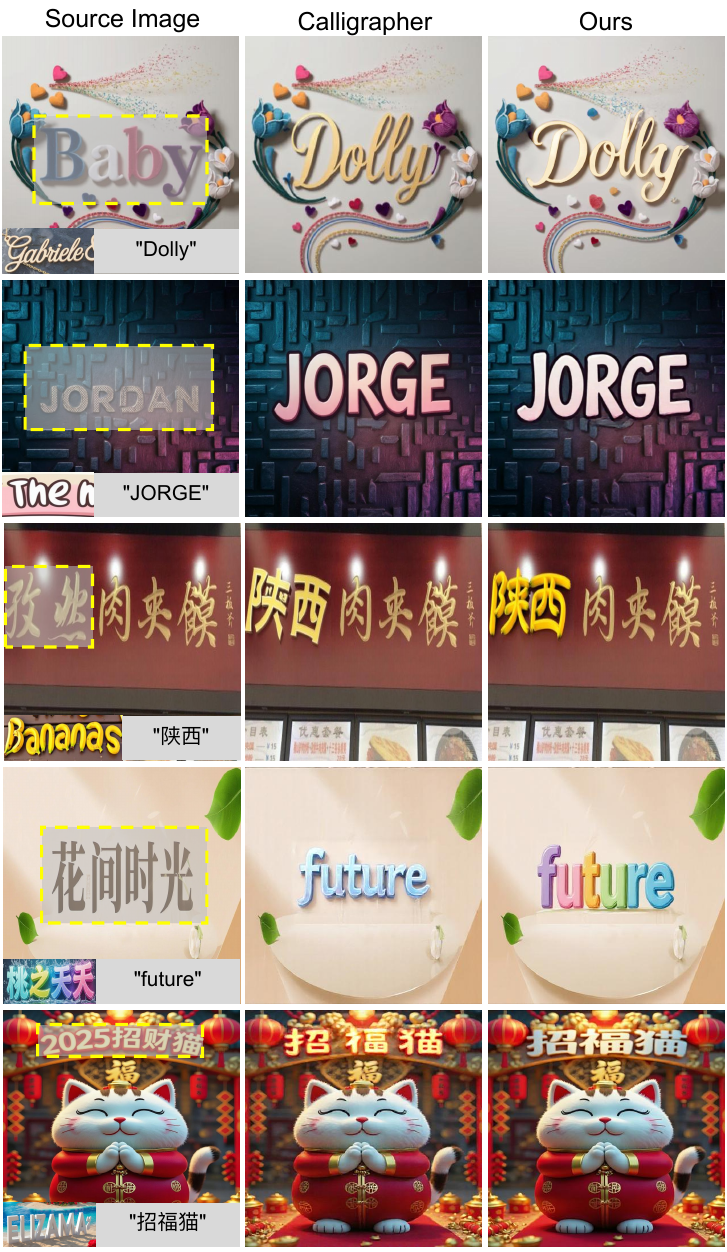}
   \caption{Qualitative comparison on the StyleText-CE benchmark under the external-style reference setting.}
   \label{fig:qua2}
\end{figure}

\section{Experiments}
\subsection{Dataset}
\noindent\textbf{Data synthesis.}
Inspired by the ArtText Diffusion model from POSTA~\cite{chen2025posta}, which uses an inpainting-based approach to generate stylized text through character-level masks and style prompts, we create a bilingual stylized dataset for model training. We first collect 77 bilingual TrueType font files that support both Chinese and English characters to generate corresponding text masks. Next, we prepare 2,218 artistic style prompts and 10,000 text-free background images. These resources serve as input to the ArtText Diffusion model to synthesize the bilingual style dataset.

\noindent\textbf{Training data.}
For the training data, we begin with 10,000 high-quality bilingual images accompanied by text box pairs as a base dataset. We then follow the data construction pipeline introduced in Calligrapher~\cite{ma2025calligrapher} to generate 2,000 stylized English samples, utilizing the Flux model along with explicit style text prompts. Finally, we select and incorporate an additional 2,000 high-quality synthesized bilingual style images into the final training set.
\subsection{Benchmark}
\noindent\textbf{AnyWord-Eval.}
AnyWord-Eval~\cite{tuo2023anytext} consists of 1,000 English samples and 1,000 Chinese samples, provides a balanced bilingual testbed.

\noindent\textbf{StyleText-CE.}
To further evaluate the model’s style control capability in cross-lingual scenarios, we construct StyleText-CE, containing 2021 high-quality Chinese and English scene text images. The benchmark supports evaluation under both self-style and external-style reference modes: in the self-style mode, the style reference image is the original image itself, whereas in the external-style mode, the style pattern is provided by an extra reference image. These two settings allow for comprehensive assessment under both monolingual and cross-lingual configurations.

\subsection{Implementation Details}

\noindent\textbf{Model Initialization.}
For Dual-Branch Style Encoder, the Textual Style Branch adopts InternViT~\cite{chen2024internvl} as a text-oriented encoder, initialized with the pretrained weights of TokenFD~\cite{guan2025token}, which has been trained on vision--language alignment tasks involving visual text and demonstrates strong bilingual text perception capabilities. 
We further perform style-preserving text segmentation training on our synthesized bilingual style dataset to enhance its ability to localize text regions and capture text style attributes. 
The Visual Prior Branch uses the text-insensitive generic visual encoder SigLIP~\cite{tschannen2025siglip} to provide robust visual representations and avoid shortcuts in the self-reference mode.

Our DIT model is initialized with the weights of TextFlux~\cite{xie2025textflux}, a multilingual scene text generation model that demonstrates strong multilingual generation capability and is built upon FLUX.1-Fill-Dev~\cite{flux2024}.

\noindent\textbf{Bilingual Text Segmentation Model.}
During the computation of the Text Style Consistency Loss and the inference-time style injection, we further fine-tune Hi-SAM\cite{hi-sam_tpami24} on our synthesized bilingual style dataset to enhance its segmentation performance on multilingual text images.

\noindent\textbf{Training Setup.}
During training, we freeze all parameters of the DIT model and the encoder components of both branches in the Dual-Branch Style Encoder. 
Only the remaining components of the two branches and the linear layers that map the text-aware style embedding to the style key--value pairs are optimized, preserving the representational capacity of the pretrained modules.
The model is trained at a resolution of $512 \times 512$, using the AdamW~\cite{adamw2019} optimizer with a learning rate of $2\times10^{-5}$, a batch size of 2, and 16-step gradient accumulation.

\subsection{Quantitative Results}
For text correctness evaluation, we adopt two widely used metrics: (i) Sen.Acc, which measures line-level text recognition accuracy, and (ii) NED, the normalized edit distance between the generated text and the ground truth. Following ~\cite{tuo2023anytext}, we employ the same OCR tool ~\cite{chen_ppocrv3_corr22} to extract textual content from generated images and compute Sen.Acc.
For style similarity assessment, we use: (i) FID, which quantifies the distribution-level discrepancy between visual style features, and (ii) LPIPS, which measures perceptual similarity at the sample level.

We first conduct quantitative analysis on AnyWord-Eval benchmark under the editing (self-style reference) paradigm, as shown in Table~\ref{tab:model_comparison}.
Since Calligrapher does not originally support Chinese text generation, we adopt the TextFlux backbone only for the Chinese evaluation, while keeping its original configuration for the English setting.
Our method outperforms existing approaches in both text accuracy and style similarity. 
Specifically, the sentence accuracy of our model exceeds TextFlux by \textbf{5.6} and \textbf{3.4} percentage points on English and Chinese data, respectively.
The normalized edit distance also improves consistently, increasing from 0.8018 to 0.8575 on English and from 0.7934 to 0.8187 on Chinese.
In terms of style similarity, our method shows strong style adherence, with FID scores reduced by \textbf{16.7\%} on English data and \textbf{8.21\%} on Chinese data compared with TextFlux.
Moreover, our model achieves the lowest LPIPS scores among all methods, decreasing from 0.187 to 0.161 on English and from 0.204 to 0.187 on Chinese, indicating that the generated results are perceptually closer to the reference styles.

To comprehensively evaluate the ability of style-conditioned scene text generation, we further conduct experiments on the StyleText-CE benchmark. Since Calligrapher is the only existing method that supports the external style reference paradigm, we compare our model with it under this setting.

For evaluation on StyleText-CE, the conventional computation of FID is not applicable in the external style reference scenario. Therefore, we employ a bilingual text segmentation model to extract text regions from both the reference and generated images, and compute FID and LPIPS on the segmented regions to avoid background interference.

As shown in Table~\ref{tab:styletext_ce_results}, our method consistently outperforms Calligrapher across all four metrics. 
Under the self-style reference setting, our model improves Sen.Acc and NED by 0.11 and 0.095, while reducing FID and LPIPS by approximately 9.8 and 0.025, respectively. 
Under the external-style reference setting, the gains remain clear: Sen.Acc and NED increase by 0.135 and 0.11, while FID and LPIPS decrease by about 15.1 and 0.028.

It is worth noting that FID and LPIPS values are generally higher in the cross-lingual configurations of the self-style reference setting and in the external style reference setting. This is because style similarity is evaluated only on text regions, where the glyph shapes of the reference text and the generated text often differ significantly, leading to larger perceptual and distribution-level discrepancies.

\subsection{Qualitative Results}
In the qualitative analysis, we first compare our method with TextFlux\cite{xie2025textflux} and Calligrapher\cite{ma2025calligrapher} under the self-style reference paradigm (noting that TextFlux does not support reference-based generation—its output style is solely influenced by the background context, and is included here only for visual comparison). As shown in Figure~\ref{fig:qua1}, the first two rows present monolingual cases, where our method produces outputs with minimal color deviation from the style reference and achieves noticeably higher style consistency. The last two rows illustrate cross-lingual scenarios, in which our model demonstrates robust style feature extraction and transfer across languages, generating text with higher fidelity and more consistent stylistic alignment.
For the external style reference paradigm, Figure~\ref{fig:qua2} presents a comparison between our approach and Calligrapher. From the cases in the first and fifth rows, we observe that our Dual-Branch Style Encoder effectively focuses on text regions, enabling more accurate extraction of style information even under visually cluttered backgrounds. The other examples further show that our method produces more consistent styles, generating text that matches the reference style more closely.

\subsection{Ablation Study}

\begin{table}[t]
\centering
\small
\setlength{\tabcolsep}{5pt}
\caption{Quantitative ablation results. We analyze the contributions of the Dual-Branch Style Encoder, the Text Style Consistency Loss ($\mathcal{L}_{tsc}$), and the inference-time style injection module. ``text. branch'' and ``visual. branch'' denote the textual style branch and the visual prior branch of the encoder, respectively.}
\begin{tabular}{lcccc}
\toprule
\textbf{Variant} & \textbf{Sen.Acc $\uparrow$} & \textbf{NED $\uparrow$} & \textbf{FID $\downarrow$} & \textbf{LPIPS $\downarrow$} \\
\midrule
w/o text. branch & 0.618 & 0.775 & 133.62 & 0.536 \\
w/o visual. branch & 0.634 & 0.789 & 124.18 & 0.503 \\
\midrule
w/o $\mathcal{L}_{tsc}$ & 0.629 & 0.788 & 126.94 & 0.509 \\
w/o infer. style inj. & 0.646 & 0.796 & 118.36 & 0.494 \\
\midrule
\textbf{Full (Ours)} & \textbf{0.659} & \textbf{0.804} & \textbf{113.47} & \textbf{0.482} \\
\bottomrule
\end{tabular}
\label{tab:ablation_styletext}
\end{table}

In this section, we analyze the contributions of the key components in StyleTextGen. 
Table~\ref{tab:ablation_styletext} reports quantitative results on the StyleText-CE benchmark. We first analyze the Dual-Branch Style Encoder. Removing either branch results in clear degradation in both text accuracy and style similarity. In particular, removing the textual style branch increases the FID from 113.47 to 133.62, highlighting the importance of accurately extracting text-specific style cues. Removing the visual prior branch also degrades performance, raising the FID to 124.18, which indicates that global visual context provides complementary information for style modeling. Similar trends are observed for Sen.Acc, NED, and LPIPS.
We further evaluate the Text Style Consistency Loss and the inference-time style injection strategy. Removing $\mathcal{L}_{tsc}$ increases the FID to 126.94, demonstrating its role in enforcing alignment between the generated text and the reference style. 
Disabling the inference-time style injection also degrades performance, increasing the FID to 118.36. These results indicate that both components contribute to improved text fidelity and visual style consistency.

Figure~\ref{fig:ab} provides qualitative comparisons for the ablation study. 
The left part shows the effects of removing the Text Style Consistency Loss and the inference-time style injection module. 
Without the text style consistency loss, the generated text tends to inherit visual cues from surrounding backgrounds rather than from the reference text, leading to inconsistent style transfer. 
Introducing the loss improves style alignment, while the inference-time style injection further refines visual coherence and produces more stable results.
The right part illustrates the ablation of the Dual-Branch Style Encoder. 
Removing the textual style branch significantly weakens style similarity, particularly under complex backgrounds where accurate extraction of text-specific style cues is required. 
In contrast, removing the visual prior branch mainly affects glyph shape and layout fidelity. 
These observations highlight the complementary roles of the two branches in capturing both local textual style and global visual context.
\section{Conclusion}

In this work, we introduced StyleTextGen, a framework for style-conditioned multilingual scene text generation. StyleTextGen employs a dual-branch style encoder that combines a textual style branch with a visual prior branch to learn robust, text-aware style representations across different scripts. To promote consistent text styling, we introduce a text style consistency loss that enforces uniform style within text regions, and an inference-time style injection strategy that further refines style alignment between reference and generated text. We also construct StyleText-CE, a benchmark for evaluating style-conditioned scene text generation. Experiments on AnyWord-Eval and StyleText-CE demonstrate that StyleTextGen improves both text accuracy and style similarity over existing methods, and exhibits strong cross-lingual generalization.
\newpage
\section*{Acknowledgements} This work is supported by the National Natural Science Foundation of China (Grant NO 62376266 and 62406318).

{
    \small
    \bibliographystyle{ieeenat_fullname}
    \bibliography{main}
}



\end{document}